\def\year{2018}\relax
\documentclass[letterpaper]{article} 
\usepackage{aaai18}  
\usepackage{times}  
\usepackage{helvet}  
\usepackage{courier}  
\usepackage{url}  
\usepackage{graphicx}  
\usepackage{color}
\usepackage{tabularx,threeparttable}
\usepackage{multirow}
\usepackage{caption}
\usepackage{bbm}
\usepackage{array}
\usepackage{enumitem}
\usepackage{amsmath, amssymb,bm}
\usepackage{booktabs}
\usepackage{graphicx,epstopdf,subfigure}
\usepackage{multirow}
\usepackage{tabularx,threeparttable}
\usepackage{array}
\usepackage{xcolor}
\usepackage{comment}
\usepackage[T1]{fontenc}
\usepackage[utf8]{inputenc}
\usepackage{url}
\usepackage{mathtools}
\usepackage{bbm}
\usepackage{threeparttable}
\usepackage{bm}
\begin{document}
\title{SFCN-OPI: Detection and Fine-grained Classification of Nuclei Using \\ Sibling FCN with Objectness Prior Interaction}
\author{Yanning Zhou$^{1}$, Qi Dou$^{1}$, Hao Chen$^{1}$, Jing Qin$^{2}$, Pheng-Ann Heng$^{1}$
        \\ $^{1}$Department of Computer Science and Engineering, The Chinese University of Hong Kong
        \\ $^{2}$Centre for Smart Health, School of Nursing, The Hong Kong Polytechnic University
        \\ \texttt {\{ynzhou,qdou,hchen,pheng\}@cse.cuhk.edu.hk, harry.qin@polyu.edu.hk}}

  \maketitle

\begin{abstract}

Cell nuclei detection and fine-grained classification have been fundamental yet challenging problems in histopathology image analysis.
Due to the nuclei tiny size, significant inter-/intra-class variances, as well as the inferior image quality, previous automated methods would easily suffer from limited accuracy and robustness.
In the meanwhile, existing approaches usually deal with these two tasks independently, which would neglect the close relatedness of them. In this paper, we present a novel method of sibling fully convolutional network with prior objectness interaction (called SFCN-OPI) to tackle the two tasks simultaneously and interactively using a unified end-to-end framework.
Specifically, the sibling FCN branches share features in earlier layers while holding respective higher layers for specific tasks. More importantly, the detection branch outputs the objectness prior which dynamically interacts with the fine-grained classification sibling branch during the training and testing processes. With this mechanism, the fine-grained classification successfully focuses on regions with high confidence of nuclei existence and outputs the conditional probability, which in turn benefits the detection through back propagation. Extensive experiments on  colon cancer histology images have validated the effectiveness of our proposed SFCN-OPI and our method has outperformed the state-of-the-art methods by a large margin.

\end{abstract}

\section{Introduction}
In digital histopathology image analysis, cell nuclei detection and fine-grained classification are crucial prerequisites for cellular morphology processing, such as computation of size, texture, shape, as well as other imagenomics~\cite{xing2016robust}, and furthermore assisting the cancer malignancy diagnosis~\cite{hamilton2000classification}.
Take the colon cancer, commonly originating from the glandular epithelial cells, as an example, recognizing the epithelial cells and assessing their morphologic changes are important for grading the cancer levels and providing guidance for therapeutic procedures~\cite{fleming2012colorectal,ricci2007identification}.
Given that manual nuclei detection is extremely time-consuming and suffers from high inter-observer variance, researchers have been dedicated to exploring automatic methods to efficiently and accurately detect the nuclei from histopathology images and conduct further analysis such as classifying the detected nuclei into fine-grained sub-categories~\cite{park2013segmentation,xie2017efficient}.

\begin{figure}[t]
	\centering
	\includegraphics[width=0.47\textwidth]{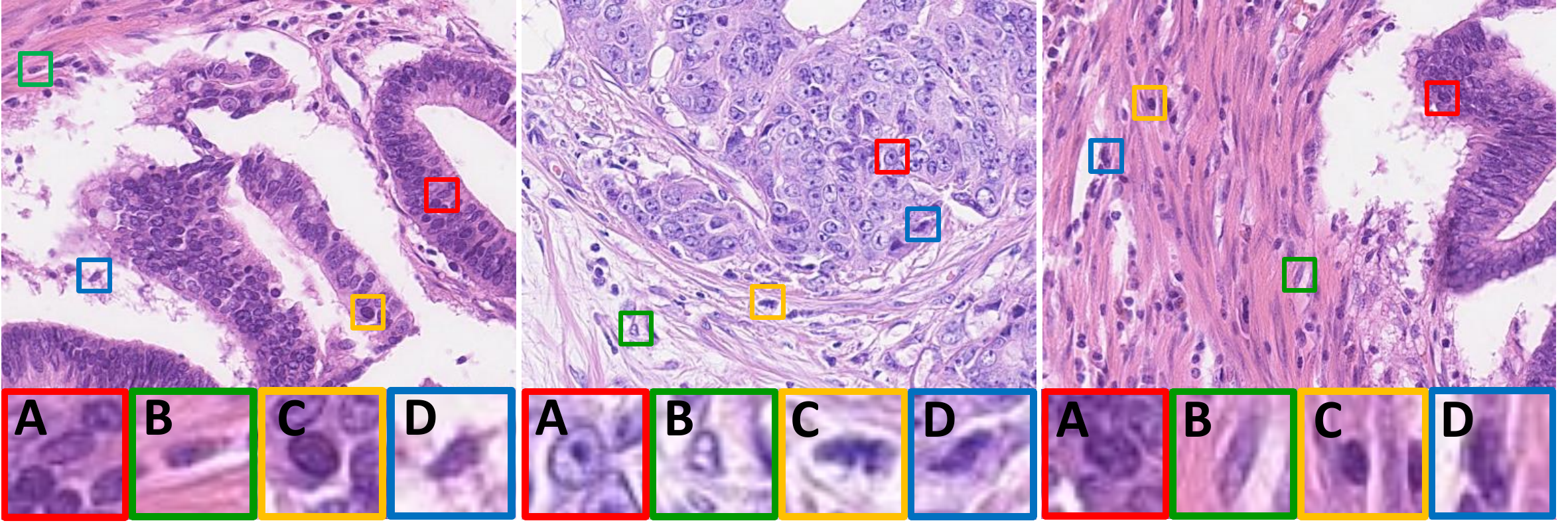}
	\caption{Examples of the histopathology images for nuclei detection and fine-grained classification. For clearer illustration, different sub-categories of the nuclei are zoomed in with the larger boxes, i.e., A. epithelial nuclei, B. fibroblast nuclei, C. inflammatory nuclei, D. miscellaneous nuclei.
 }
	\label{fig:challenge}

\end{figure}

However, development of automatic digital histopathology image analysis approach is difficult.
Firstly, the size of nuclei is quite small in histopathology images, with radius usually only a few pixels.
Secondly, sub-categories among the nuclei have significant inter-/intra-class variances in their shape and chromatin texture, which are related to different histopathology grades.
Thirdly, the tumorigenic cells tend to clutter together and hold erratic morphology changes, which would lead to complicated contexts with degenerated glandular structures and hence bring obstacles to cell nuclei identification.
These challenges are shown in Fig.~\ref{fig:challenge}.
In addition, the presence of inferior image quality caused by poor staining, unfocused photoing or failed digital sampling further increases the difficulty of automatic nuclei detection and fine-grained classification.

Early nuclei analysis methods relied on handcrafted features based on the intensity~\cite{wang2007novel,li2010multiple} as well as morphology~\cite{al2010improved,park2013segmentation}.
Unfortunately, these low-level features would suffer from limited representation capability and robustness, leading to inaccurate detections especially on hard cases such as densely adhesive cells and malignant cells with highly heterogeneous shape~\cite{xing2016robust}.

Recent revolution of the deep convolutional neural network (CNN) has achieved great success on object detection~\cite{ross2015fast,Shaoqing2015faster} in natural images.
The state-of-the-art methods usually first suppress the majority of non-object regions and generate sparse region of interest (ROI) proposals, and then classify them into predefined categories and regress a refined bounding box for each detected object.
Though these methods have set promising performance on natural images, they are hardly applicable to histopathology images. This is because that the cell populations in a histopathology image would vary widely, ranging from several to hundreds.
In addition, the nuclei size is quite small and the nuclei are only annotated with centroid points.
Hence, solutions based on ROI proposal and bounding box regression would be infeasible in our application.
Many attempts of adapting CNNs on histopathology images have also been explored~\cite{cirecsan2013mitosis,cruz2014automatic,Chen_2016_CVPR,chen2016mitosis}.
Particularly, for nuclei analysis,~\citeauthor{xie2017efficient} (2017) proposed an efficient and robust model based on the fully convolutional network (FCN) and set outstanding nuclei detection performance on four histopathology datasets.
The method of~\citeauthor{sirinukunwattana2016locality} (2016) employed two sequential CNNs to separately detect and classify the nuclei in patch-based manner.
Though achieved promising performance, this practice would detach the intrinsic association between the two tasks and unnecessarily involve redundant computations.

In this paper, we propose to solve both tasks of nuclei detection and fine-grained classification with a unified framework which is learned in an end-to-end manner.
Specifically, we design an efficient FCN architecture consisted of two sibling branches (referred to as SFCN).
They share features in earlier layers and hold respective higher layers targeting for each specific task.
More importantly, we propose a novel objectness prior interaction (OPI) mechanism which dynamically interacts the sibling branches during learning process.
In this manner, our model sufficiently captures the close relatedness between the sub-tasks and hence boosts the accuracies of both nuclei detection and fine-grained classification.
Our main contributions are summarized as follows:
\begin{itemize}
\item We propose a novel architecture, i.e., SFCN-OPI, which is able to simultaneously detect the cell nuclei and classify them into sub-categories at high efficiency and accuracy.
\item Our proposed OPI is able to effectively invoke interaction between the highly correlated sub-tasks and make our model to respect inter-task relatedness. Moreover, the end-to-end learning which jointly optimizes the sibling branches enables the nuclei detection and fine-grained classification to benefit from each other.
\item Extensive experiments have validated effectiveness of the proposed SFCN-OPI, exceeding the state-of-the-art methods by a large margin.
\end{itemize}

\section{Related work}
Automatic nuclei detection from the histopathology image is defined as acquiring the location of nuclei without depicting the accurate boundaries~\cite{xing2016robust}.
It is regarded as a prerequisite step to narrow down the interesting areas for important following analysis including fine-grained classification of the nuclei into sub-categories.

Early studies on automatic nuclei detection and classification can be mainly divided into groups of gradient and intensity based method, morphology based method and machine learning based method.
Gradient and intensity methods took advantages of Euclidean distance and gradient magnitude map to select feature points in local area~\cite{wang2007novel,li2010multiple}.
In contrast, morphology based methods employed more complex geometric features to detect certain structures of the nuclei.
For examples,~\citeauthor{park2013segmentation} (2013) proposed a modified ultimate erosion method to separate the overlapping convex objects relying on the contour and shape descriptors.
\citeauthor{al2010improved} (2010) designed a multi-scale Laplacian of Gaussian filter with spatial constraints to identify the nuclei.
Later on, machine learning based methods aimed to build pixel/patch-wise classifiers such as SVM, random forest, and probabilistic models~\cite{mualla2013automatic,Khan2013A,sommer2012learning}, trained with comprehensive hand-crafted features including local intensities, gradients, global textures and shape information.
However, all these traditional methods relied on low-level features with limited representation capability and tended to be sensitive to cell morphology changes especially for tumorous cells.

\begin{figure*}[t]
	\centering
	\includegraphics[width=1.0\textwidth]{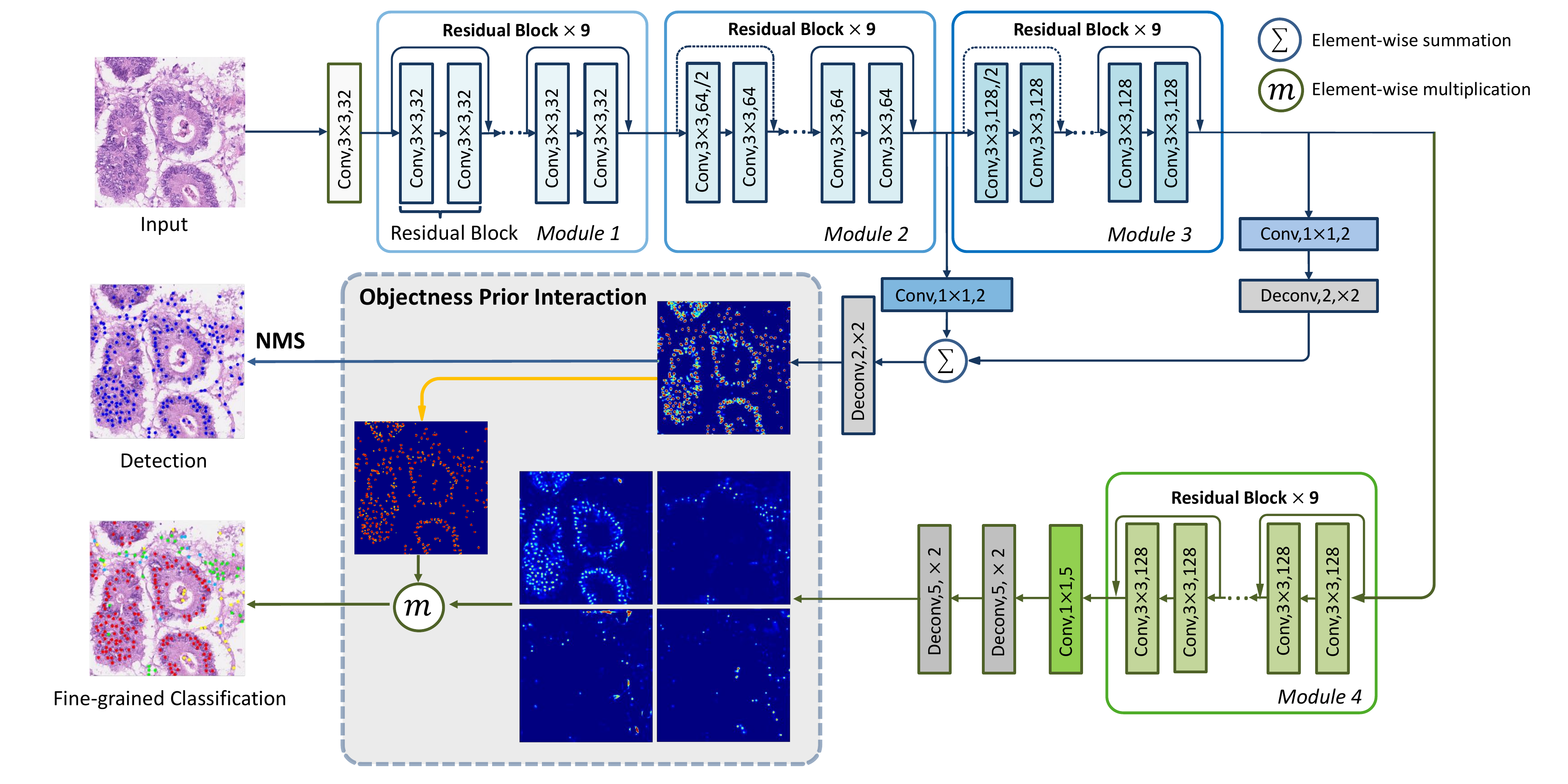}
	\caption{ An overview of our proposed SFCN-OPI for nuclei detection and fine-grained classification in a unified framework. }
	\label{fig:overview}
\end{figure*}
Recently, with the advancements of feature learning, deep CNNs have been employed for histopathology image analysis to a large extent.
\citeauthor{cirecsan2013mitosis} (2013) proposed to detect the mitosis by classifying the cropped small patches with a CNN in sliding window manner.
Later on, \citeauthor{chen2016mitosis} (2016) proposed a cascade framework, with an FCN architecture to first rapidly screen the candidates and another CNN to discriminate the true positives.
For nuclei analysis, \citeauthor{xie2015beyond} (2015) proposed  the convolutional neural network based structured regression
mode called SR-CNN, which replaced the last CNN layer to structure regression layer and designed a proximity mask to encode topological structure information, achieving an accurate detection of the nuclei from histopathology images.
Based on SR-CNN,~\citeauthor{sirinukunwattana2016locality} (2016) proposed a spatially constrained convolutional neural network (SC-CNN) which ameliorated SR-CNN with spatial constrained layer to detect the nuclei.
They also built another independent CNN to further classify the detected nuclei into four sub-categories.
This work has achieved outstanding performance on a typical histopathology image dataset, and hence we regard it as the current state-of-the-art method for nuclei detection and classification. However, this method used patch-based inference which would suffer from the inferior efficiency. More importantly, it detached the highly correlated tasks of nuclei detection and fine-grained classification and treated them in separate networks without taking advantages of their mutual information.

\section{Method}
\subsection{FCN Architecture with Sibling Branches}

Different from the previous methods which employed patch-based training and sliding window based testing,
we propose to utilize the fully convolutional network to efficiently and accurately compute the nuclei in histopathology images.
Considering that the sub-tasks of nuclei detection and fine-grained classification are highly correlated, we design our FCN architecture with sibling branches that shares low-level features in early layers while each aiming for a particular task at respective higher layers.
To meet the challenges of the problem mainly due to the complicated morphological variances of nuclei, we propose a 80-layer deep FCN, and its detailed configurations are illustrated in Fig.~\ref{fig:overview}.

To effectively train the deep FCN, we introduce the residual connections~\cite{he2016deep} into our network for benefiting the optimization process.
In each residual block $x_{i+1} = \mathcal{W}_s x_i + \mathcal{F}_i(x_i; \{\mathcal{W}_s\})$, we utilize a stack of two $3\!\times\! 3$ convolutional layers followed by batch normalization (BN) and ReLU nonlinearity to conduct the residual function $\mathcal{F}_i$.
The $\mathcal{W}_s$ is used to downsample feature maps by $1\!\times\! 1$ convolution with a stride of 2, otherwise, the $\mathcal{W}_s x_i$ is identity mapping.

Let $H$ and $W$ respectively represent the height and width of the input image, it initially passes one $3\!\times\!3$ convolutional layer with 32 filters followed by a BN and ReLU layer.
Next, we hierarchically stack three modules each consists of nine residual blocks to learn representations of the histopathology image.
In the beginning of modules 2 and 3, we employ a strided convolution to reduce the resolution while doubling the number of the feature maps.
Specifically, the sizes of feature maps are ($H$, $W$), ($H/2$, $W/2$), ($H/4$, $W/4$) and the numbers of feature maps are 32, 64, 128 in modules 1, 2, 3.

In higher layers, the network construction splits into sibling branches, one for nuclei detection and the other for fine-grained classification.
For the detection branch, we connect the shared layers to a $1\times 1$ convolutional layer with two filters to obtain the highly abstract representations.
Towards accurate detection of the nuclei region, we fuse these representations with features abstracted after the second module with element-wise summation, which helps the network to take advantages of the important multi-scale information. Finally, the fused feature maps are deconvolved into original size as the input image and forwarded to a Softmax layer so that we obtain the objectness probability prediction for each location in the histopathology image.

For the fine-grained classification branch, the network requires richer high-level information to distinguish across the sub-categories. In this regard, we add another module with 9 residual blocks following the shared components.
In module 4 in Fig.~\ref{fig:overview}, we apply a $1\times 1$ convolutional layer with 5 filters to convert the high dimension features into five feature maps.
After the deconvolution and Softmax operations, we obtain five score maps, corresponding to four nuclei sub-categories and the background.
Noting that the sibling branches have interactions during training and prediction mode, which improves the performance dramatically. Details about the interactions shall be described in following subsections.

\subsection{Classification with Combined Objectness Prior}

Previous state-of-the-art method used to firstly detect the nuclei with a CNN and then classify the detections into sub-categories with a separately trained network~\cite{sirinukunwattana2016locality}.
Actually, the two tasks are highly correlated, especially given that each sub-category in the classification is indeed a subset of the detection results.
Forcefully detaching the two sub-tasks would not only involve superfluous computations but also prohibit the mutual benefits that can be acquired from each other.
On the other hand, the conventional multi-task wise network design where different tasks purely share weights in lower layers while respectively holding independent higher layers is also not an optimal solution.
In these regards, we propose to resolve both problems using a unified framework with dynamic interactions between two sibling branches.

However, simultaneously localizing and recognizing sub-categories of the nuclei under a unified framework is quite challenging.
Some challenging difficulties for the fine-grained classification would come from the significant inter-/intra-class variances (see Fig.~\ref{fig:challenge}) and the severe class imbalance of the nuclei.
More importantly, learning to distinguish sub-categories of the nuclei should be particularly focused on the detected objects, while excluding those massive background areas.

In this paper, we propose the objectness prior interaction mechanism to elegantly tackle the simultaneous detection and fine-grained classification of nuclei in histopathology images.
The core insight lies in how to obtain the classification probability by interacting the sibling FCN.
Specifically, the probability indicating objectness, output from the detection branch, is transferred as an informative prior to the fine-grained classification branch.
In this way, the fine-grained classification branch actually learns a conditional probability given the confidence of objectness at each location.
In other words, this branch only needs to particularly focus on distinguishing different nuclei sub-categories without being bothered by noisy background which can be successfully suppressed by the objectness prior.
As a result, the fine-grained classification probability is acquired in an interactive manner, by combining the objectness prior and the conditional probability obtained from sibling FCN branches.

Formally, for each location in the input image, the detection branch outputs the scores $p^{det} \! = \! \{ p^{bkg}, p^{obj}\}$ where the $p^{obj}$ indicates its probability of being the nuclei.
In the meanwhile, the fine-grained classification branch predicts the conditional probability $p^{cls|obj} \! = \! \{ p_0^{cls|obj},p_1^{cls|obj},...,p_K^{cls|obj}\}$, representing the confidence of each location being assigned to each sub-category given the objectness prior $p^{obj}$.
The $K$ denotes the number of sub-categories and is 4 in our setting.
Finally, we calculate the fine-grained classification probability for each location as follows:
\begin{equation}
\label{eq:cls_prob}
p^{cls=k}= p^{obj} \cdot p_k^{cls|obj},
\end{equation}
where $k \in [0,1,...,K]$ denotes a particular class and $k=0$ is the background.
The $p^{cls=k}$ is the classification probability of the location belonging to sub-category class $k$.

\subsection{Joint Loss Function of SFCN-OPI}
Towards our target to jointly learn the tasks of nuclei detection and fine-grained classification,
the loss function is composed of both detection binary classification errors and fine-grained multi-class classification errors.

Specifically, the detection error for each location is negative log-likelihood of the predicted probability.
To meet the situation of class imbalance between the nuclei and background regions, we introduce class weights into the loss:
\begin{equation}
\label{eq:loss_det}
\small
\mathcal{L}_{det} \! = \! -\frac{1}{N}  \sum_{i=1}^N \! \left( \mathbbm{1}(y_i \! = \! 0) \log p_i^{bkg} + \alpha \mathbbm{1}(y_i \! = \! 1) \log p_i^{obj} \right)
\end{equation}
where $y_i$ is the ground truth detection label of pixel $i$, with $y_i \! = \! 0$ being background and $y_i \! = \! 1$ being the nuclei.
The $p_i^{bkg}$ and $p_i^{obj}$ respectively present the predicted score of being background and nuclei at pixel $i$, and they are obtained via Softmax function and summed up to 1. The $N$ is the total number of equivalent training samples during learning.
The $\alpha$ is weighted-loss hyper-parameter calculated according to the proportion of positive and negative pixels in the image.

For the fine-grained classification, this branch should ideally focus only on the nuclei regions and learn to assign them into sub-categories.
With our attempt to tackle detection and classification in a unified FCN framework, particular strategies have to be proposed to exclude the massive background regions when learning the classification branch.
To achieve this, we propose to suppress the loss at background regions with low confidence of objectness.
Specifically, we transform the $p^{obj}$ into a binary mask by thresholding it, and use the mask as a gate when calculating the fine-grained classification loss.

For each pixel $j$, we calculate its classification error only if its objectness prior $p_j^{obj}$ exceeds a predefined probability threshold $t_p=0.8$.
In this way, the fine-grained classification branch is particularly focused on the high-confidence nuclei samples, disregarding the background regions which have been successfully addressed by the detection branch.
Formally, at pixel $j$, we calculate its fine-grained classification negative log-likelihood loss as follows:
\begin{equation}
\label{eq:loss_cls}
\small
\mathcal{L}^j_{cls} =  - \gamma_{c_j} \mathbbm{1}(p_j^{obj} \! > \! t_p) \cdot \log p_j^{cls=c_j}
\end{equation}
where $c_j$ is its ground truth class-specific label, and $p_j^{cls=c_j}$ is the predicted probability towards this sub-category.
The $\mathbbm{1}(p_j^{obj} \! > \! t_p)$ is the indicator function serving as the gate.
Similar to calculation of the detection errors, we also adopt weighted loss for the classification branch to balance across the sub-categories, and $\gamma_{c_j}$ is the corresponding weighted loss multiplier for class $c_j \in [0,1,...,K]$.

Overall, by representing the parameters in the network with $\mathcal{W}$, we jointly optimize our network with sibling FCN by minimizing the following loss function:
\begin{equation}
\label{eq:loss_overall}
\small
\mathcal{L} = \mathcal{L}_{det} + \lambda \frac{1}{N_{cls}} \sum_j \mathcal{L}^j_{cls} + \beta ||\mathcal{W}||^2_2,
\end{equation}
where the first term is the detection loss, the second term is the fine-grained classification loss, and the third term is the weight decay.
The $N_{cls}$ is the number of pixels that are counted by the objectness gate mask in calculation of the classification loss. The $\lambda$ and $\beta$ are hyper-parameters to balance the three loss components.

We optimize our SFCN-OPI framework using the standard back-propagation~\cite{rumelhart1988learning}.
With the objectness prior interaction mechanism as described in Eq.~(\ref{eq:cls_prob}), the gradients derived from the fine-grained classification loss can flow to those layers of the detection branch, and vice versa.
In other words, those poor detection outputs with inaccurate objectness probability can be punished not only via the detection loss but also via the gradients from the classification branch.
With this manner, the close relatedness between both sub-tasks are sufficiently taken advantages of by the dynamic interactions during the joint learning process.

\subsection{Training Strategies and Testing Inference}

To train the sibling FCN, we first need to construct two ground truth masks which respectively annotate the regions of nuclei and their sub-categories.
The provided annotations of the dataset are single points indicating the centroids of the nuclei and the class-specific labels of the nuclei.
In practice, we mark the positive region as a circle area centering at the centroids of the nuclei with a small radius.
This supervision mask design is simple yet effective, and benefits the network to gain translation invariance during the learning process.
Some previous works designed promaxity map where the centroids of the nuclei are given higher value than the nuclei edges~\cite{xie2017efficient}.
Their supervision masks to train FCN are more carefully designed, but their practice could only employ regression loss which is not optimal for classification, especially multi-class classification tasks.

Training strategy of the SFCN-OPI needs to be carefully designed, since that the fine-grained classification branch deeply interacts with the objectness prior obtained from the detection branch.
In other words, effective learning of the fine-grained classification branch should be conditioned under high-quality objectness prior heatmap.
In this regard, we divide the training process into three stages: 1) pre-training of the nuclei detection branch; 2) pre-training of the classification branch with the shared and fixed detection layers; 3) updating the entire network jointly.
The sibling branches are well established and they dynamically interact with each other via the OPI mechanism.
In this way, the performance, both of detection and fine-grained classification, can be further boosted by effectively capturing the inter-task relatedness.
Our training strategy to learn the SFCN-OPI has been experimentally validated and we shall present observation results in the following section.

During the testing inference, given an input image, we obtain two scoremaps (background/nuclei) from the detection branch and five scoremaps from the fine-grained classification (background/ epithelial/ fibroblast/ inflammatory/ miscellaneous~nuclei) branch.
To obtain the detection results, we conduct the non-maximum suppression (NMS) on the detection scoremap to obtain the points of nuclei.
When getting the fine-grained classification results, the objectness priors are element-wisely multiplied onto the classification scoremaps.
Each detected location is then classified into the sub-category which has the highest probability across the five scoremaps.

\section{Experiment}

In order to evaluate the performance of the proposed network, we conduct extensive experiments on a typical cell nuclei dataset\footnote{https://www2.warwick.ac.uk/fac/sci/dcs/research/tia/data}~\cite{sirinukunwattana2016locality}, which contains $100$ hematoxylin and eosin (H\&E) stained histopathology images of colorectal adenocarcinoma with totally $29756$ nuclei annotated at their centers for detection task.
Out of these nuclei, there are $22444$ nuclei further labeled into four sub-categories for the classification task with $34.3\%$ of them epithelial nuclei, $31.1\%$ inflammatory nuclei, $25.5\%$ fibroblast nuclei and $9.1\%$ miscellaneous nuclei.
As the dataset just provides the coordinates of cell centroids, in order to formulate the training label form of the proposed FCN-based network, we employ a small circle mask centered at each centroid with $3$-pixel radius as the ground truth when training.

\subsection{Implementation Details}

In training phase, we performed data augmentation to further enlarge the training data.
Specifically, we cropped original images into $64\times 64$ sub-images and randomly combined zooming, rotating, shearing, horizontal/vertical flipping and channel shifting.
We used Xavier uniform initializer~\cite{glorot2010understanding} to initialize all weights and set initial bias as zero.
We used stochastic gradient descent with Nesterov momentum as the optimizer, with a batch size of $200$, the momentum of $0.9$, and weighted decay of $0.0001$.
We set the initial learning rate as $0.01$, and decayed it to $0.001$ at $100$ epochs and $0.0001$ for the next $50$ epochs.
We trained our model with a NVIDIA Titan Xp GPU which took about $6$ hours for training convergence.
The $100$ histopathology images were randomly divided into training, validation, and test sets at a ratio of 7: 1: 2.

\subsection{Evaluation Metrics}
We harness the evaluation metrics used in~\citeauthor{sirinukunwattana2016locality} (2016) to validate our experimental results and facilitate comparison.
As for detection, we use precision, recall and F1 score to evaluate the performance of various methods.
Same as~\citeauthor{sirinukunwattana2016locality} (2016), we consider the prediction results located in a circle centered at a cell centroid with $6$-pixel radius as true positive (TP).
All detected points outside these regions are considered as false positive (FP).
Each mask region corresponded to an annotated point which does not contain any detected point is considered as a false negative (FN).
As for classification, we use weighted average of the precision, recall and F1 score of each class of nuclei to evaluate the performance of various methods. The weights are the percentages of different classes in the total labeled nuclei (please refer to the first paragraph of this section).

\subsection{Experimental Results}

\textbf{Effectiveness of objectness prior interaction.}
The challenge to combining nuclei detection and fine-grained classification is how to leverage the relatedness between two tasks to acquire more distinguishing features for classification, where large inter-/intra-class variances and massive noisy background exist. In order to assess the effectiveness of the proposed training schemes, particularly the OPI mechanism, we first conduct a set of ablation experiments.
We employ three networks in these experiments.
1) We train a single FCN to directly classify each pixel of the image into 5 classes (4 nuclei sub-categories and the background), called FCN-5CLS. It basically utilizes the same network architecture as the SFCN-OPI, with the only difference being that the FCN-5CLS has no sibling branches. Its detection performance is calculated by summing up the probability maps of the four sub-categories of nuclei. Weighted loss is also introduced to facilitate training and the multiplier for each class is obtained according to the pixel proportions of different labels.
2) We train a network which has two FCN branches for detection and classification respectively, but there is no OPI involved during the training (SFCN). The classification loss of SFCN is calculated only for the pixels close to the ground truth.
3) The last one is our proposed SFCN-OPI with both sibling branches and OPI (Ours in Table~\ref{tab:Compared results}).

In Table~\ref{tab:Compared results}, the FCN-5CLS yields lowest F1 score for both detection ($0.790$) and classification ($0.298$).
This is because the large inter-/intra-class variances and the massive background regions increase the difficulty to acquire discriminative representations of the nuclei for accurate classification.

 \begin{table}[t]
  \centering
 \caption{Experimental results of ablation analysis, different training strategies of our method and comparison with other approaches.}
 \small
 \label{tab:Compared results}
 \begin{tabularx}{0.46\textwidth}{p{0.68in}p{0.23in}p{0.23in}p{0.23in}p{0.23in}p{0.23in}p{0.23in}}
 \toprule
 \multicolumn{1}{l}{\multirow{2}{*}{Methods}}  & \multicolumn{3}{c}{Detection}  & \multicolumn{3}{c}{Classification} \\
 \cmidrule(r){2-4}\cmidrule(l){5-7}
\multicolumn{1}{c}{}              & ~~~P            &~~~ R      & \multicolumn{1}{c}{F1} & ~~~P    & ~~~R& ~ F1 \\
 \midrule
 FCN-5CLS&0.741& 0.867&0.790&0.466&0.264&0.298\\
 SFCN& 0.784&0.844 &0.807&0.450&0.561&0.496 \\
 \midrule
 SFCN-OPI-1& 0.764 &\textbf{0.890}  &0.816 &0.573&0.667& 0.613 \\
 SFCN-OPI-2&0.788  &0.885  & 0.828 &0.674&0.759& 0.711 \\
 \midrule
 SSAE& 0.617 & 0.644 &0.630 &~~~~-&~~~~-&~~~~ - \\
 LIPSyM& 0.725 & 0.517 & 0.604 &~~~~-&~~~~-&~~~~ - \\
 CRImage& 0.657 & 0.461 & 0.542 &~~~~-&~~~~-&~~~~ - \\

 SR-CNN& 0.783 & 0.804 & 0.793 &~~~~-&~~~~-& 0.683 \\
 SC-CNN& 0.781 & 0.823 & 0.802 &~~~~-&~~~~-& 0.692 \\
  \midrule
\textbf{Ours}& \textbf{0.819} & 0.874 &\textbf{0.834} &\textbf{0.718}&\textbf{0.774}& \textbf{0.742} \\
\bottomrule

 \end{tabularx}

 \footnotesize{Note: the - means the results were not reported by that method.}
 \end{table}

  \begin{figure*}[htbp]

    \centering

     \includegraphics[width=1.0\textwidth]{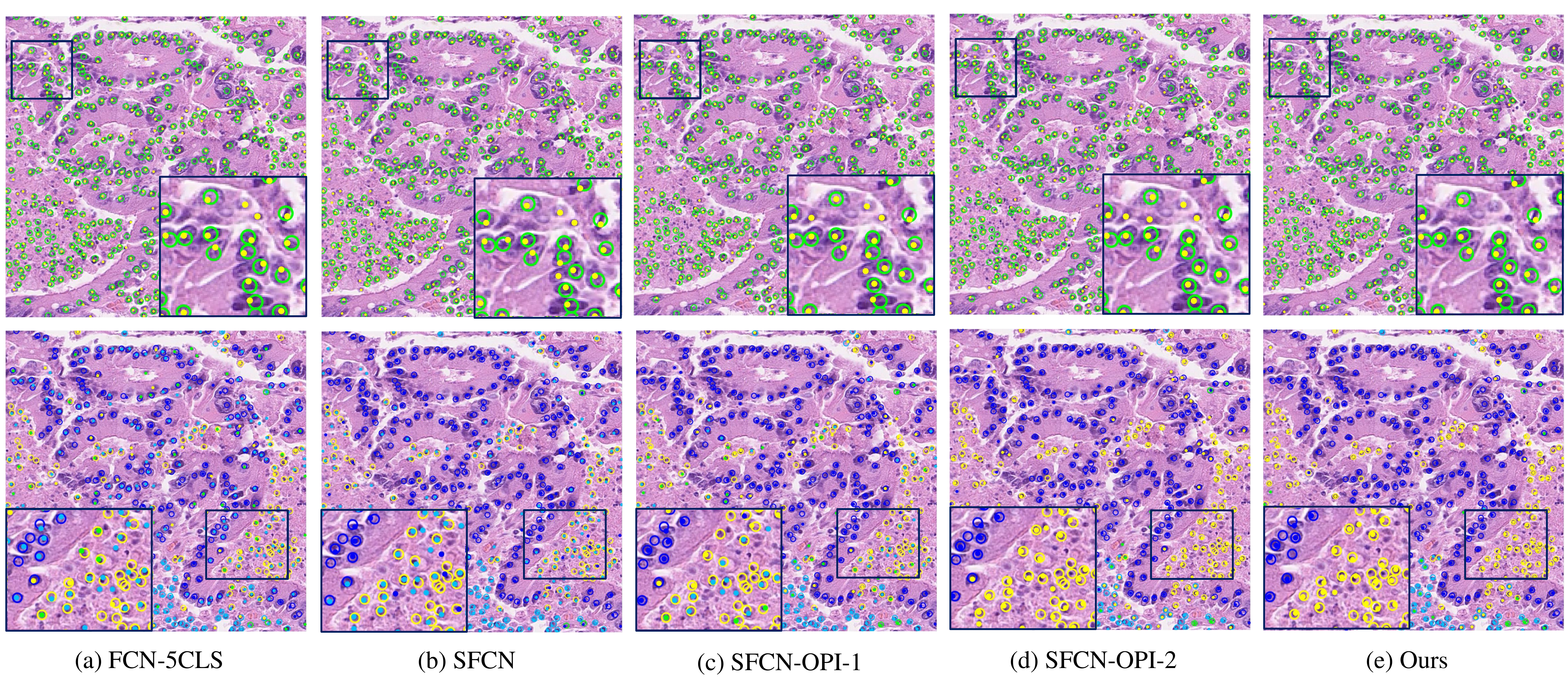}
    \caption{Typical results of our nuclei detection (first row) and fine-grained classification (second row). In the first row, yellow dots represent our predicted results and the green circles indicate the ground-truth. In the second row, dots and circles in different colors indicate different sub-categories (navy blue: epithelial nuclei, green: fibroblast nuclei, light blue: inflammatory nuclei, yellow: miscellaneous nuclei).}

    \label{fig:Compared}
  \end{figure*}

Compared with FCN-5CLS, the SFCN achieves striking improvement of the F1 score for classification task, boosting it from $0.298$ to $0.496$. This is because the classification loss is calculated only for the pixels close to the ground truth, which weakens the massive background influence and forces the classification branch to focus on distinguishing the inter-class variance among nuclei sub-categories.
The the detection F1 also slightly improves from $0.790$ to $0.807$.
The underlying reason is that, the nuclei detection task and fine-grained classification task are highly correlated.
In this regard, sharing weights in early layers between two FCN branches can benefit the training of both branches.

Compared with FCN-5CLS and SFCN, the results of our proposed SFCN-OPI achieve significant improvement, attaining F1 score of $0.834$ and $0.742$ for detection and classification, respectively.
Rather than SFCN which directly applies the regions near ground truth to calculate the classification loss, SFCN-OPI dynamically chooses locations with high objectness probability to calculate the classification loss. The locations with high objectness probability vary in different training iterations, so that wherever there exist false positive regions in detection branch, there would also receive extra penalties from the classification loss.
By dynamically selecting high-confidence nuclei regions to take into account in the classification loss, the proposed SFCN-OPI effectively suppresses hard false positives as well as massive noise of large background regions. It learns more accurate objectness prior probability, which in turn contributes to learning more precise conditional classification probability.

\noindent\textbf{Experiments of different training strategies.}
We further test the effectiveness of different training strategies on the proposed SFCN-OPI.
To obtain a comprehensive insight on the interaction of the two branches, we train our network in three different training strategies:
1) initially train detection branch, and then freeze the detection branch and train the fine-grained classification branch (SFCN-OPI-1);
2) train detection branch firstly, and then jointly train the whole network (SFCN-OPI-2);
3) train detection branch firstly, next freeze detection branch and train classification branch, and finally jointly train the whole network (Ours).

In Table~\ref{tab:Compared results}, compared with the SFCN-OPI-1, the SFCN-OPI-2 improves the F1 score of classification from $0.613$ to $0.711$. The difference between these two strategies is that the latter setting has a joint training step, which allows the shared weights in early layers to be optimized by the losses of both branches through the back-propagation procedure.
Moreover, our proposed strategy further improves the F1 score of classification to a large extent (achieving 0.742).
The difference between SFCN-OPI-2 and ours is that the latter has a seperete classification pre-training step, which gives the classification branch a relatively better initialization before joint training. Results validate that balancing the learning difficulties of sub-tasks to a comparable level helps to sufficiently leverage the benefit of OPI mechanism.

\noindent\textbf{Comparison with other methods.}
We then compare our SFCN-OPI with five well-established or state-of-the-art methods on the same dataset.
We compare our detection results with above methods and further compare our classification results with those of two state-of-the-art methods, which deal with, albeit separately, both detection and classification tasks.
Three of them are based on stacked autoencoder or hand-crafted features~\cite{xu2016stacked,kuse2011local,yuan2012quantitative} while other two are deep CNN based methods~\cite{sirinukunwattana2016locality,xie2015beyond}.
We employ the results of these methods on the same dataset reported in~\cite{sirinukunwattana2016locality} for the comparison.
Specifically,~\citeauthor{xu2016stacked} (2016) harnessed unsupervised stacked sparse autoencoder (SSAE) while
~\citeauthor{kuse2011local} (2011) and~\citeauthor{yuan2012quantitative} (2012) leveraged a local isotropic phase symmetry measure (LIPSyM) and a hierarchical multiresolution model (CRImage) for detection task, respectively.
~\citeauthor{xie2015beyond} (2015) developed a CNN based structure regression model (SR-CNN) and~\citeauthor{sirinukunwattana2016locality} (2016) proposed a spatially constrained convolutional neural network (SC-CNN) based on SR-CNN, which regresses the likelihood of a pixel being the nuclei based on constrains of spatial prior and then training another CNN independently to classify nuclei into sub-categories after detection.
The SC-CNN performed both detection and classification tasks and achieved state-of-the-art results.

It is observed that the three CNN-based approaches, including ours, achieve much higher F1 scores for detection than the approaches based on hand-crafted features or stacked autoencoder, demonstrating the features learned from deep CNNs are more discriminative than hand-crafted features and the features learned from shallow unsupervised networks.
Furthermore, our approach achieves the highest F1 score in both sub-tasks, which corroborates the effectiveness of leveraging highly correlated information by the objectness prior interaction.
Our approach also outperforms the SR-CNN and SC-CNN in classification task by a large margin, improving the F1 score from $0.692$ to $0.742$.
Note that our SFCN-OPI-2 also achieves better classification result than that of both SR-CNN and SC-CNN, further demonstrates the advantages of joint training with OPI over separate training of detection and classification networks.

\noindent\textbf{Qualitative results.}
We further provide some typical results of detection and classification, as shown in Fig.~\ref{fig:Compared}.
The first row in Fig.~\ref{fig:Compared} shows detection results of different networks with various training scheme.
The detected nuclei are marked by yellow dots and the ground truths are denoted as green circles.
The magnified region enclosed with a black box clearly shows that our method successfully eliminates some hard false positives that cannot be well tackled by other networks.
This is because that these hard samples receive stronger penalty not only from the detection branch but also from the fine-grained classification branch through the joint loss function.
The second row shows fine-grained classification results where we employ different colors to denote different nuclei sub-categories.
The magnified region highlights the results of epithelial and miscellaneous nuclei (marked by navy blue and yellow), which are significant for colon cancer diagnosis but have high degree of pleomorphism and quite challenging to be accurately identified.
Our method achieves an obvious improvement of identification compared with other networks.
It illustrates that the OPI mechanism can greatly improve the performance of fine-grained classification by extensively exploiting the relatedness of the detection and classifications tasks.

\section{Conclusion}
In this paper, we propose a novel framework, i.e, SFCN-OPI, aiming to conduct both cell nuclei detection and fine-grained classification using a unified architecture.
Our designed network consists of two sibling branches, with each outputting predictions for a specific sub-task.
To sufficiently take advantages of the relatedness between the detection and fine-grained classification tasks, we dynamically interact the two branches during the learning process by transferring the objectness prior from nuclei detection to classification.
The joint learning of the entire framework enables both sub-tasks to enjoy mutual benefits from each other and hence improve their prediction accuracies.
Extensive experiments validate the efficacy of our method and the effective training strategies in practice.
Experimental results on colon glandular cancer dataset demonstrates the outstanding performance of our SFCN-OPI, exceeding the state-of-the-art approaches by a significant margin.
Last but not least, our proposed framework, as well as the spirit of respecting the relatedness in the multi-tasks network design, is general and can be extendable to many other histopathology image analysis problems.

\section{Acknowledgments}
This project is supported by the Research Grants Council of the Hong Kong Special Administrative Region (Project No. CUHK 14225616) and the following grants from Innovation and Technology Fund of Hong Kong (Project no. ITS/041/16 and ITS/304/16).

\begin{quote}

    \bibliographystyle{aaai}
    \bibliography{bibfile1}

\end{quote}

\end{document}